%% file: acl_latex.tex
\pdfoutput=1

\documentclass[11pt]{article}

\usepackage[]{acl}

\usepackage{soul}

\usepackage{color}
\usepackage{xcolor}
\definecolor{darkgreen}{RGB}{83,129,53}
\definecolor{darkred}{RGB}{163,21,21}

\usepackage{tikz}
\usepackage{subfig}
\usepackage{tcolorbox}

\usepackage{pgfplots}
\usepackage{wrapfig}
\usepackage{amssymb,mathrsfs}
\usepackage{amssymb}
\usepackage{array}
\usepackage{pgfplotstable}
\usepackage{pgf}
\usepackage[normalem]{ulem}
\usepackage{colortbl}
\usepackage{ulem}
\usepackage{xspace}

\usepackage{listings}

\usepackage{scalerel} 
\usepackage{times}
\usepackage{latexsym}
\usepackage{graphicx}
\usepackage{amsmath}
\usepackage{multirow}
\usepackage{amsthm}
\usepackage{amsfonts}
\usepackage{bm}
\usepackage{booktabs}
\usepackage{algorithm}
\usepackage{algpseudocode}
\usepackage{verbatim}
\usepackage{tabularx}

\usepackage{times}
\usepackage{latexsym}

\usepackage[T1]{fontenc}

\usepackage[utf8]{inputenc}

\usepackage{microtype}

\usepackage{inconsolata}

\usepackage{graphicx}

\usepackage{multirow}
\usepackage{booktabs}
\usepackage{adjustbox}

\usepackage{booktabs}   
\usepackage{threeparttable} 
\usepackage{graphicx}   
\usepackage{xcolor}     

\usepackage{stfloats} 

\title{Towards Robust Universal Information Extraction: Dataset, Evaluation, and Solution}

\author{
 \textbf{Jizhao Zhu\textsuperscript{1,2}},
 \textbf{Akang Shi\textsuperscript{2}},
 \textbf{Zixuan Li\textsuperscript{1}}$^{*}$,
 \textbf{Long Bai\textsuperscript{1}},
 \textbf{Xiaolong Jin\textsuperscript{1}}$^{*}$,\\
 \textbf{Jiafeng Guo\textsuperscript{1}},
 \textbf{Xueqi Cheng\textsuperscript{1}}
\\
 \textsuperscript{1}Key Laboratory of Network Data Science and Technology,\\
 Institute of Computing Technology, Chinese Academy of Sciences \\
 \textsuperscript{2}School of Computer Science, Shenyang Aerospace University, Shenyang, China\\
 \texttt{\{zhujz@sau.edu.cn, shiakang@stu.sau.edu.cn,lizixuan@ict.ac.cn}
}

\newcommand \footnoteONLYtext[1]
{
	\let \mybackup \thefootnote
	\let \thefootnote \relax
	\footnotetext{#1}
	\let \thefootnote \mybackup
	\let \mybackup \imareallyundefinedcommand
}
\begin{document}

\maketitle
\footnoteONLYtext{$^{*}$Corresponding authors.}
\begin{abstract}
In this paper, we aim to enhance the robustness of Universal Information Extraction (UIE) by introducing a new benchmark dataset, a comprehensive evaluation, and a feasible solution. Existing robust benchmark datasets have two key limitations: 1) They generate only a limited range of perturbations for a single Information Extraction (IE) task, which fails to evaluate the robustness of UIE models effectively; 2) They rely on small models or handcrafted rules to generate perturbations, often resulting in unnatural adversarial examples. Considering the powerful generation capabilities of Large Language Models (LLMs), we introduce a new benchmark dataset for Robust UIE, called RUIE-Bench, which utilizes LLMs to generate more diverse and realistic perturbations across different IE tasks. Based on this dataset, we comprehensively evaluate existing UIE models and reveal that both LLM-based models and other models suffer from significant performance drops. To improve robustness and reduce training costs, we propose a data-augmentation solution that dynamically selects hard samples for iterative training based on the model's inference loss. Experimental results show that training with only \textbf{15\%} of the data leads to an average \textbf{7.5\%} relative performance improvement across three IE tasks.

\end{abstract}

\section{Introduction}

Information Extraction (IE) aims to extract structured knowledge from unstructured text based on predefined types of entities, relations, and events. Universal Information Extraction (UIE), which seeks to unify the extraction of various knowledge types through a single model, has achieved significant progress in recent years. Most existing studies have primarily focused on enhancing the overall performance of UIE models, typically evaluated on fixed test sets. However, they often overlook the robustness (and generalization ability)  of UIE models, which are crucial when handling real-world text.

To measure the robustness of IE models, some studies focus on constructing benchmark datasets by generating adversarial examples with small perturbations. For example, RockNER~\cite{lin2021rockner} employs a rule-based approach and BERT~\cite{bert} to generate two kinds of perturbations for Named Entity Recognition (NER); ~\citet{li2021robustness} generates adversarial examples for Relation Extraction (RE) by random replacing words with synonyms or similar words generated by some Natural Language Processing (NLP) tools. \citet{liu-etal-2020-context} replaces verbs and context using similar words generated by GloVe~\cite{glove} for Event Detection (ED). Overall, existing benchmark datasets typically have two limitations: 1) They generate limited kinds of perturbations for individual IE tasks, making it difficult to comprehensively evaluate the robustness on UIE models across various IE tasks; 2) They generate adversarial examples typically using small models or handcrafted rules, often resulting in unnatural samples.

Considering the powerful NLP capabilities of Large Language Models (LLMs), we leverage them in this paper to generate more diverse and realistic perturbations. After human verification of the annotation accuracy of LLMs, we obtain a new benchmark dataset for Robust UIE, called RUIE-Bench. RUIE-Bench contains \textbf{11580} samples and includes \textbf{14} distinct kinds of perturbation across three mainstream IE tasks, i.e., NER, RE, and ED. Based on RUIE-Bench, we conduct a comprehensive evaluation of existing UIE models, including 8 open-source LLMs, 4 closed-source LLMs, 4 traditional IE models, and 4 fine-tuned UIE models. We obtain some intriguing observations from the experimental results, such as open-source LLMs have a significant performance gap compared with closed-source LLMs in both original tests and perturbation tests. LLM-based UIE models demonstrate better robustness than traditional IE models under certain perturbations. However, both types of models suffer from significant performance drops.

To improve the robustness of UIE models, a common solution is data augmentation. Since the training cost is also a key factor, especially for LLM-based UIE models, we further propose a \textbf{L}oss-guided \textbf{D}ata \textbf{A}ugmentation (LDA) solution to enhance the robustness of models using a limited number of samples. Specifically, we first generate additional adversarial examples for training. Then, the inference loss on these samples is leveraged to dynamically select the most challenging ones to fine-tune the UIE model. Using the fine-tuned model, we iteratively calculate the inference loss and select hard samples for the next round of training. The experimental results demonstrate that training KnowCoder~\cite{knowcoder} with just 15\% of the augmented data using LDA yields a \textbf{7.5\%} relative improvement in average performance on RUIE-Bench, compared with the state-of-the-art models. This performance is comparable to the fully trained model using the entire augmented dataset. Additionally, when evaluated on the unseen dataset, KnowCoder with LDA outperforms the fully trained model by an average of \textbf{8.9\%} across three IE tasks.

In summary, our contributions are as follows:
\begin{itemize}
  \item We construct RUIE-bench, which contains \textbf{11580} samples with \textbf{14} distinct perturbations generated by LLMs across various IE tasks, which is the most comprehensive benchmark dataset with the most diverse perturbations for robust UIE. 
  \item Based on the RUIE bench, we comprehensively evaluate existing IE models. The evaluation results highlight that current IE models exhibit robustness issues against perturbations. 
  \item To improve the robustness with limited samples, we further propose a loss-guided data augmentation solution, which achieves performance comparable to training with the full dataset by using only 15\% of the data. Moreover, when evaluated on unseen datasets, LDA outperforms the fully trained model with 8.9\% F1 on average across three IE tasks.

\end{itemize}

\section{Related Work}
\textbf{Universal Information Extraction} can be classifier into two kinds of methods, classification-based~\cite{oneie, nguyen-etal-2022-learning} and generation-based UIE methods~\cite{UIE, InstructUIE, knowcoder}. The former mainly adopts the end-to-end joint extraction mode, enhancing cross-task interactions with global dependency modeling for unified extraction. The latter aims to generate structural information rather than extracting structural information from plain text. Recently, some UIE methods employ various types of prompts to enable LLMs to understand schemas and extract the corresponding knowledge. For example, InstructUIE~\cite{InstructUIE} applies a text-style prompt for schema understanding. In contrast, KnowCoder~\cite{knowcoder} uses a code-style prompt to transform UIE into code generation, achieving state-of-the-art performance.

\noindent \textbf{Robustness Research for IE} primarily focuses on constructing benchmark datasets for individual IE tasks. For NER, RockNER~\cite{lin2021rockner} replaces the original entities with entities of relevant types from Wikidata\footnote{\url{https://www.wikidata.org}}, and employs BERT~\cite{bert} to substitute context words; ~\citet{jin-etal-2023-adversarial} adopts disentanglement and word attribution methods to identify keywords and injects character-level perturbations for these words; \citet{srinivasan-vajjala-2023-multilingual} applies multiple perturbations, including random entity replacement, Bert-based contextual modifications, and paraphrase generation. For RE, ~\citet{li2021robustness} generates adversarial examples by randomizing word substitutions using RoBERTa~\cite{roberta} or using synonyms; ~\citet{nolano-etal-2024-pointing} generates adversarial examples by replacing entities in triples using various strategies, including same-type and different-type substitutions. For ED, \citet{liu-etal-2020-context} uses GloVe~\cite{glove} to replace similar words to generate adversarial examples.

To enhance the robustness of IE, the existing method~\cite{lin2021rockner} expands the original training data by generating additional examples. Furthermore, ~\citet{li2021robustness, jin-etal-2023-adversarial, srinivasan-vajjala-2023-multilingual} employ perturbation techniques to generate adversarial examples for training. While these methods typically require a large number of adversarial examples, none focus on improving model robustness using only a small number of augmented samples.

\section{The Construction of RUIE-Bench}

In this section, we first introduce the methods of utilizing LLMs to generate adversarial examples. Subsequently, we provide the concrete construction process 
of the RUIE-Bench dataset.

\subsection{LLM-based Adversarial Example Generation}\label{perturbation methods} In UIE, adversarial examples refer to samples intentionally perturbed to mislead UIE models while maintaining their original semantics or appearance. Mathematically, given an input sample $s$ with its corresponding label $y$, an adversarial example $s'$ is generated by applying a small perturbation, which is constrained to ensure that it remains close to the input space. The goal of adversarial examples is to cause a model to predict an incorrect label.

\begin{figure*}[tbp]  
 \centering
 \includegraphics[width=1.0\textwidth]{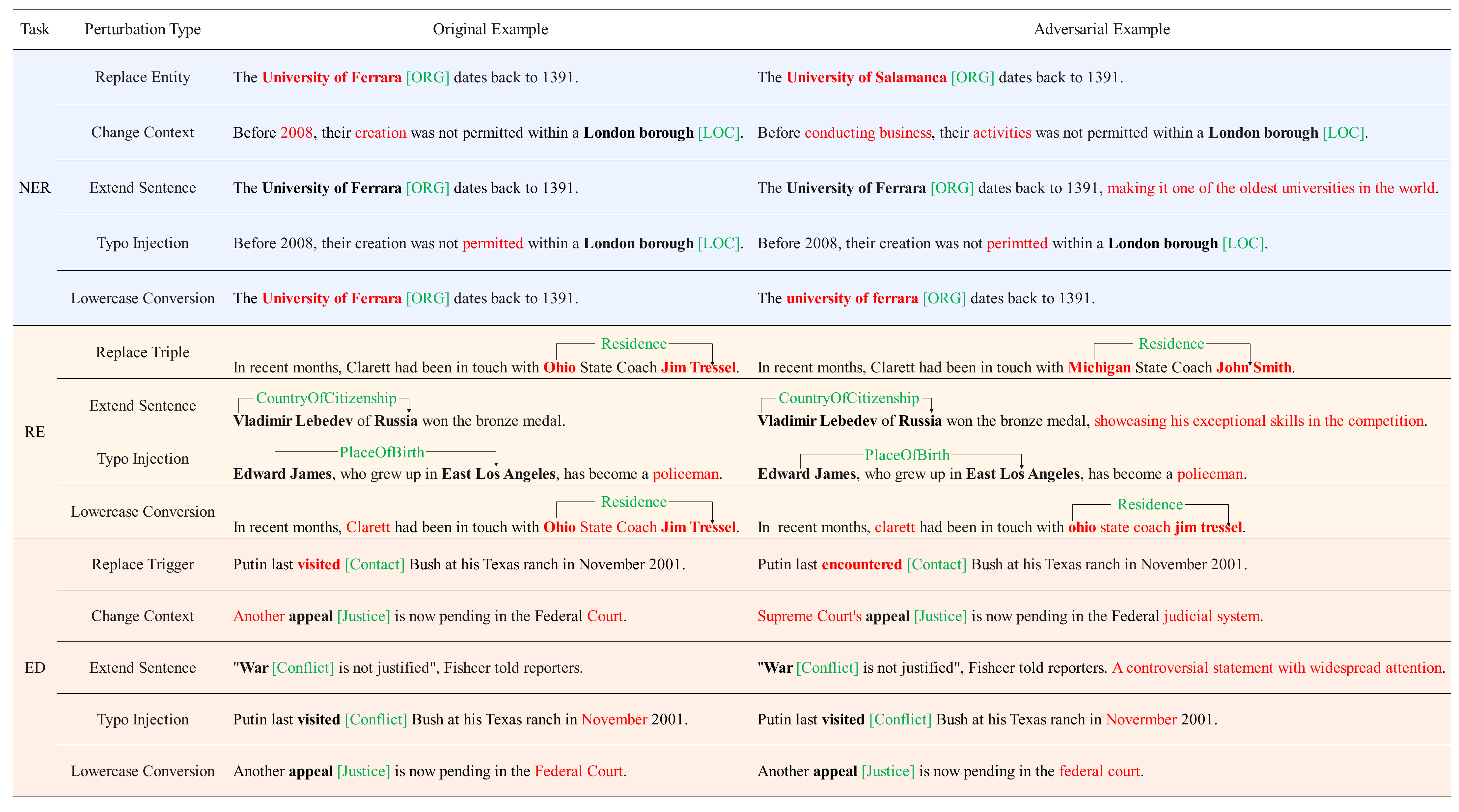}
 \caption{Illustration of generated adversarial examples with different kinds of perturbations.}
 \vspace{-4mm}
  \label{fig:noise type example}
\end{figure*}

To generate more diverse and realistic adversarial examples, we employed LLMs to simulate different kinds of perturbations for NER, RE, and ED tasks by designing various prompts for different IE tasks. Additionally, we utilized two general rule-based perturbations, which were applied across all IE tasks. A comprehensive demonstration of these adversarial examples is presented in Figure~\ref{fig:noise type example}. In what follows, we will introduce these different kinds of perturbations for IE tasks in detail.

\paragraph{Replace Entity, Triple, and Trigger.} 
A robust UIE model should be able to identify entities, relations, and events based on the context of the sentence corresponding to the task rather than memorizing each entity, relational triple, or event trigger and their corresponding types. To prevent the model from memorizing specific patterns instead of reasoning based on context, we introduce perturbations to existing sentences. 

Such perturbations need to ensure the consistency of the original type during replacement. Although previous studies~\cite{lin2021rockner, nolano-etal-2024-pointing} have introduced similar perturbations in NER and RE samples, such as replacing entities based on rules, these methods may lead to incorrect replacement. Moreover, for ED samples, replacing triggers while ensuring the type remains unchanged is an extremely challenging task. Given this, we instruct GPT-4~\cite{GPT4} to replace entities, relational triples, or event triggers while preserving their types and ensuring that other content remains unchanged. The corresponding prompts are provided in Appendix~\ref{appendix:prompt1}.

\paragraph{Change Context.} 

A robust UIE model should keep its performance even when the contextual content of the sample changes due to various perturbations. To evaluate the robustness of the UIE model against contextual variations, we introduce perturbations to context words. It should be noted that this method is only used for samples of NER and ED tasks since altering context words in RE may disrupt the semantic relations between entity pairs.

In the previous methods, the mask language model BERT~\cite{bert} is used for changing context, but these methods often generate semantically inaccurate or syntactically incorrect words, and only one word can be generated for a single mask position. Therefore, we use LLMs to change the context in sentences. Specifically, we first get rid of punctuation, entities, event triggers, and stop words in sentences, leaving only meaningful context words. And randomly choose up to four words to replace with the [MASK] token. Then, 
GPT-4~\cite{GPT4} is instructed to generate three predictions for each [MASK] token and randomly select one for replacement. The prompts can be found in Appendix~\ref{appendix:prompt1}.

\paragraph{Extend Sentence.} 
Generally, a robust UIE model is capable of accurately extracting the required information, even in the face of complex sentence structures or long text situations. In order to better evaluate the robustness of the UIE model in handling complex sentences or long text situations, we enhance the semantic depth of the sentences by adding semantically relevant content (such as contextual details, historical facts, or explanatory clauses), thereby increasing the complexity of the sentences.

Prior research has not explored the robustness of IE models under similar perturbations. Considering that added content must maintain semantic coherence and meaningfulness, we employ LLMs to implement this perturbation. For specific tasks, we instruct GPT-4~\cite{GPT4} to adhere to corresponding constraints. For NER, new sentences must preserve the original entity boundaries and types without introducing new entities. For RE, the original relational triples must remain unchanged, with no additional relational information introduced. For ED, the original event triggers in sentences must be retained without incorporating new event information. The prompts are provided in Appendix~\ref{appendix:prompt1}.

\paragraph{Typo Injection.} 
In reality, typos are common. However, a robust UIE model should continue to make accurate predictions when dealing with these typos. To simulate these common text spelling mistakes, we introduce typo injection. Namely, spelling mistakes are added to the words prone to errors. Initially, we tried using LLMs to inject, but it often introduced unrealistic errors. Therefore, we switched to a rule-based approach to achieve this.

We focus on sentences with more than eight words and select longer words (over six characters), as longer sentences and words are more prone to spelling errors. Additionally, we filter out stop words and high-frequency vocabulary to avoid selecting common words. We randomly choose 1-3 words from the remaining words for typo injection and avoid changing the first character for each selected word and introduce errors by randomly replacing characters, deleting characters, inserting characters, or swapping adjacent characters.

\paragraph{Lowercase Conversion.} 
Consider that in practice, users overwhelmingly spell in lowercase, especially in informal environments such as social media, email, or search queries. Therefore, lowercase conversion is used to simulate non-standard input, which helps evaluate the robustness of the UIE model in response to changes in text format.

In this method, all characters of each word are converted to lowercase, except for the first letter of the first word. This tests whether the model can still accurately extract information under non-standard input conditions, forcing it to rely on semantic understanding rather than surface features. By doing so, it not only assesses the model’s robustness but also highlights how upper and lowercase expressions affect task performance.

\begin{table*}[!htb]
\centering
\resizebox{\textwidth}{!}{%
\begin{tabular}{l|ccccc}
\toprule
\textbf{Benchmark Dataset} & \multicolumn{1}{c}{\textbf{RUIE-Bench}} & \multicolumn{1}{c}{\textbf{RockNER}} & \multicolumn{1}{c}{\textbf{DWR}} & \multicolumn{1}{c}{\textbf{Adv\_re}} & \multicolumn{1}{c}{\textbf{CSMG}} \\
 & \multicolumn{1}{c}{(Ours)} & \multicolumn{1}{c}{\citep{lin2021rockner}} & \multicolumn{1}{c}{\citep{jin-etal-2023-adversarial}} & \multicolumn{1}{c}{\citep{nolano-etal-2024-pointing}} & \multicolumn{1}{c}{\citep{liu-etal-2020-context}} \\
\midrule
\textbf{Supported tasks} & NER \& RE \& ED & NER & NER & RE & ED \\
\textbf{Covered datasets} & 6 & 1 & 3  & 1 & 2 \\
\textbf{Number of perturbation types} & 14 & 3 & 1 & 4 & 2 \\
\textbf{Number of adversarial examples} & 11580 & 13169 & - & 6,277 & - \\
\textbf{Methods for generating perturbations} & LLM \& Rule & Small model \& Rule & Small model & Small model \& Rule & Small model \\
\bottomrule
\end{tabular}%
} \caption{Comparison between RUIE-Bench and existing IE robust evaluation benchmark datasets. The - indicates that the detailed statistics of the datasets are not reported in their papers. }
\label{tab:dataset_comparison}
\end{table*}

\subsection{Dataset Construction}

To construct the RUIE-Bench dataset, we select six datasets across the three subtasks of UIE. For NER, we use the ACE05-Ent~\cite{ACE2005_DATASET}, CoNLL03~\cite{CoNLL03_Dataset}, and WikiANN~\cite{wikiann_Dataset} datasets; for RE, we select the ACE05-Rel~\cite{ACE2005_DATASET} and NYT~\cite{NYT_DATASET} datasets; for ED, we use the ACE05-Evt~\cite{ACE2005_DATASET} dataset. To balance sample sizes across these subtasks, we perform stratified sampling from the test sets of respective datasets, adhering to the principle of maintaining distributional consistency. Specifically, we conduct stratified sampling on all label types (including NULL-type) within each test set, obtaining 1,000 NER samples, 800 RE samples, and 676 ED samples. For adversarial example generation, we apply the perturbation methods detailed in Section~\ref{perturbation methods}. For each original sample, corresponding adversarial samples are generated for every perturbation type, while retaining those samples that cannot be perturbed. During generation, we implement strict quality control through manual verification: any sample containing errors is immediately discarded and regenerated until accurate samples are obtained. Through this rigorous process, we successfully construct the RUIE-Bench dataset. We present a comprehensive comparison between RUIE-Bench and existing IE robustness evaluation benchmark datasets in Table~\ref{tab:dataset_comparison}. For detailed statistics and further information regarding RUIE-Bench, please refer to Appendix~\ref{appendix:prompt2}.

\section{Loss-guided Data Augmentation}

Data augmentation is a common strategy to improve model robustness ~\cite{rebuffi2021data, wang2022toward, li2023data}. Existing methods primarily focus on synthesizing in-distribution adversarial samples for training. However, none of them focus on improving efficiency by selecting a minimal number of training samples to enhance robustness. This is especially critical for LLMs, where training costs are significantly higher due to their scale and complexity.

Inspired by ~\citet{song2023loss, buchnik2020graph, werner2023loss}, we propose a loss-guided data augmentation solution for robust UIE. The core idea is to focus on samples where the model's performance is currently suboptimal, as indicated by higher loss values, thereby potentially accelerating convergence and improving overall model performance. First, we train the initial model on the original training set and use it to compute the inference loss of the augmented samples. For each IE task, samples with high inference loss were selected to fine-tune the model. Next, based on the obtained fine-tuned model, the inference loss of the augmented samples is recalculated, followed by another round of sample selection and fine-tuning. This iterative process was repeated $t$ times until a robust model was obtained. 

Specifically, we first fine-tune the initial model $M$ using the original training data to obtain $M_0$, and employ LLMs to generate augmented data $D_\text{{aug}}$ for the original training set. During each iteration, we use $M_{t-1}$ to compute inference loss $L_i$ for each augmented sample. Then, based on selection ratio $\beta$, the samples with higher loss are selected to form a new training dataset $D_{\text{retrain}}^{(t)}$, which is subsequently used to fine-tune $M_{t-1}$ to obtain model $M_t$. After each iteration, the model is evaluated on the validation set. The algorithm terminates and returns the final model $M_t$ when the improvement on the validation set falls below the convergence threshold $\delta$. Algorithm~\ref{algorithm1} presents the training process of the proposed strategy. The details of augmented data generation are presented in Appendix ~\ref{appendix:3}.

\begin{algorithm}
 \caption{Loss-guided Data Augmentation}
  \label{alg:iterative-finetuning}
 \begin{algorithmic}[1]
 \Require Training data $D$, Initial model $M$, Selection ratio $\beta$, Convergence threshold $\delta$ \Ensure Fine-tuned model $M_t$
  
 \State Use LLMs to generate augmented data $D_{\text{aug}}$ based on $D$;
  
 \State Fine-tune $M$ on data $D$ to obtain model $M_0$;
  
 \State $t \gets 1$;

 \Repeat \For{each sample $(x_i, y_i) \in D_{\text{aug}}$} \State Compute loss: $L_i = L(\theta_{t-1}; x_i, y_i)$ \EndFor \State Sort samples in $D_{\text{aug}}$ by loss $L_i$ in descending order; \State Select top $\beta$ samples to form $D_{\text{retrain}}^{(t)}$; \State Fine-tune model $M_{t-1}$ on $D_{\text{retrain}}^{(t)}$ to obtain new model $M_{t}$; \State $\beta \gets \beta/2$; \quad  $t \gets t + 1$; \Until {performance improvement of $M_{t}$ on the validation set falls below threshold $\delta$};
  
 \State \Return $M_t$
 \end{algorithmic}
  \label{algorithm1}
\end{algorithm}

\section{Experiment Setup}
We use RUIE-Bench to evaluate the robustness of the current models, including UIE ones, traditional IE ones, and LLMs. Meanwhile, we construct an unseen dataset to measure their generalization ability.

\subsection{Metrics}
We report the span-based offset Micro F1, following previous methods~\cite{UIE, lin-etal-2020-joint}. For NER, an entity is considered correct if both its boundaries and type are accurately predicted. For RE, a relation is deemed correct if the triple, including the relation type, head entity, and tail entity, matches the gold annotations. For ED, an event trigger is considered correct if both the event type and trigger are aligned with the gold annotations.

\subsection{Training Details}
To explore effective data augmentation methods, the KnowCoder-7B-base\footnote{\url{https://huggingface.co/golaxy/KnowCoder-7B-base}} is selected as the initial model $M$. Then, we fine-tune it on the six datasets constructed for RUIE-Bench, obtaining the model $M_0$, which is denoted as $\text{KnowCoder-7B}_\text{partial}$. Additionally, the perturbation methods proposed in Section~\ref{perturbation methods} are employed to generate augmented data $D_{\text{aug}}$ based on the original training set. Next, we fine-tune the initial model using both original training data and all augmented data, obtaining the model $\text{KnowCoder-7B-Robust}$. Finally, we fine-tune the initial model using high-loss augmented samples selected according to the LDA strategy, with an initial selection ratio of 10\%. After two iterations, we successfully construct the model $\text{KnowCoder-7B-Robust}_\text{LDA}$, utilizing a total of 15\% augmented samples. 

During the fine-tuning phase, we employ LoRA~\cite{hu2021lora} for efficient parameter tuning. The LoRA rank and LoRA alpha parameters are set to 32 and 64, respectively. The learning rate is set to $3 \times 10^{-4}$, with a warm-up rate of 0.1 and a dropout rate of 0.1. The sequence length is limited to 2048 and the batch size is set to 256. Additionally, for LDA training, the selection ratio $\beta$ is set to 10\%, and the convergence threshold $\delta$ is set to 0.3, meaning that iteration stops when the  Micro F1 score improvement of the new model is less than 0.3. During validation phase, we use greedy search with a temperature of 0 and set the maximum output length to 640. All experiments are conducted on 8 x NVIDIA-A100 80G.

\subsection{Baselines}
We adopt the state-of-the-art UIE models to validate the robustness, including UIE~\cite{UIE}, IstructUIE~\cite{InstructUIE}, YAYI-UIE~\cite{YAYI-UIE}, and KnowCoder~\cite{knowcoder}. We also employ traditional IE models for robustness evaluation across the NER, RE, and ED tasks. For NER, we choose Stanza~\cite{Stanza} and TNER~\cite{TNER}; for RE, we select PFN~\cite{PFN}; and for ED, we choose EEQA~\cite{EEQA}. Additionally, we evaluate the robustness of two categories of LLMs: open-source models, including Qwen2.5-3B-Instruct, Qwen2.5-7B-Instruct, Qwen2.5-14B-Instruct\cite{qwen2.5}, Llama3-8B-Instruct~\cite{llama3},  Glm-4-9B-Chat~\cite{chatglm}, CodeLlama-7B-Instruct~\cite{codellama}, Internlm2.5-7B-Chat~\cite{internlm}, and Vicuna-7B-v1.5~\cite{vicuna}; closed-source model API services, including GPT-3.5-turbo, GPT-4-turbo~\cite{GPT4}, DeepSeek-V3~\cite{deepseekv3}, and GLM4-Plus~\cite{chatglm}. For the evaluation of LLMs, we employ the 10-shot approach to instruct LLMs to conduct IE tasks with the specific prompts provided in Appendix~\ref{appendix:5}.

\input{main_table}

\section{Results and Analyses}

\subsection{Results on the RUIE-bench dataset}
We report the  Micro F1 scores of all the models across the three IE tasks on RUIE-Bench in Table~\ref{tab:robust evaluation}. These results cover both the original test set and various perturbation settings. For the sake of space limitation in the table, we use abbreviations for these perturbations. For NER, we use P1-P5 to represent the perturbations of Replace Entity, Change Context, Extend Sentence, Typo Injection, and Lowercase Conversion, respectively. For RE, the perturbations of Replace Triple, Extend Sentence, Typo Injection, and Lowercase Conversion are denoted as P6-P9, respectively. For ED, we employ P10-P14 to represent Replace Trigger, Change Context, Extend Sentence, Typo Injection, and Lowercase Conversion, respectively. We also report the overall performance drop of all the models under the three tasks for all perturbations in the ``$\text{Drop}_\text{avg}$'' column. The robustness evaluation results of all the models are ranked, and the final ranking is shown in the ``Rank'' column. Although there are differences in the evaluation methods employed by different categories of models, we can still draw some interesting observations from the results:

(1) The models with data augmentation training show the best performance. The model $\text{KnowCoder-7B-Robust}_{\text{LDA}}$ trained with only 15\% of the augmented data using LDA achieves results comparable with $\text{KnowCoder-7B-Robust}$. It convincingly verifies the effectiveness of the proposed LDA training strategy. Furthermore, a comprehensive comparison between these two models is in Appendix~\ref{appendix:6}.

(2) LLM-based models experience relatively smaller performance drops than other models, suggesting that LLMs have stronger generalization ability. This indicates that using LLMs to improve the robustness of UIE models is a promising approach for future work.

(3) All the LLMs exhibit a significant performance drop under various perturbations, especially in the NER and RE tasks. This indicates that LLMs face serious robustness issues when dealing with UIE tasks in few-shot prompting scenarios.

(4) From the results of the Qwen models with different parameter scales, it is evident that there is a significant positive correlation between the model's parameter scale and its robustness in NER and RE tasks. In other words, an increase in the number of model parameters often accompanies an improvement in robustness. However, for ED, no similar trend is observed, as there is a relatively small performance drop and a significant performance gap between the three models.

\subsection{Results on the Unseen Dataset}

To verify whether the model trained on data with different perturbations can generalize to unseen datasets, we create an unseen dataset that does not include any samples from RUIE-Bench. Furthermore, we ensure that the types in this dataset are a subset of those in RUIE-Bench. For NER, we select OntoNotes 5.0~\citep{OntoNotes_Dataset} and random sample some instances as unseen data. Similarly, we obtain the unseen data for RE from  CoNLL04~\citep{Conll04_DATASET} and GIDS~\citep{GIDS_DATASET}. For the ED task, since no datasets with the same event types exist, we use GPT-4~\citep{GPT4} to generate 100 unseen samples, which are then manually verified for correctness.


\begin{table}
 \centering
 \resizebox{1\linewidth}{!}
 {\begin{tabular}{@{}l|ccc|c}
 \toprule
 \multirow{2}{*}{\textbf{Model}} & \multicolumn{3}{c|}{\textbf{Unseen Dataset}} & \multirow{2}{*}{\textbf{Average}} \\
 \cmidrule{2-4}
  & \textbf{NER} & \textbf{RE} & \textbf{ED} &\\ 
 \midrule
 GPT-4-turbo & 58.9 & 38.8 & 59.7 & 52.5 \\
 GLM4-Plus & 56.6 & 41.4 & 56.1 & 51.4 \\
 DeepSeek-V3 & 59.2 & 39.8 & 55.5 & 51.5 \\
  $\text{KnowCoder-7B}_\text{partial}$ & 64.9 & 40.4 & 52.2 & 52.5 \\
  $\text{KnowCoder-7B-Robust}$ & 62.8 & 47.4 & 56.6 & 55.6 \\
  $\text{KnowCoder-7B-Robust}_{\text{LDA}}$ & \textbf{67.2} & \textbf{54.6} & \textbf{60.1} & \textbf{60.6}  \\
 \bottomrule
 \end{tabular}}
 \caption{Results on the Unseen Dataset.}
  \label{tab:unseen data-results-table}
 \vspace{-4mm}
\end{table}

Table~\ref{tab:unseen data-results-table} shows the results of different models on the unseen dataset. From the table, we can find that: 1) Compared with the models without data augmentation training, the average performance of the models of $\text{KnowCoder-7B-Robust}$ and $\text{KnowCoder-7B-Robust}_{\text{LDA}}$ in the UIE tasks is significantly improved. It is justified that training with the perturbations generated in this paper can enhance the model's generalization ability. 2) It is worth noting that the $\text{KnowCoder-7B-Robust}_{\text{LDA}}$ trained with only 15\% of the augmented data using the LDA strategy achieves an average 8.9\% performance improvement compared with the $\text{KnowCoder-7B-Robust}$ using the complete set of augmented data fine-tuning. We guess that full training will lead to overfitting and thus show poor prediction ability on the unseen dataset.

\subsection{Detailed Analysis}

\begin{figure*}  
 \centering
 \includegraphics[width=0.96\textwidth]{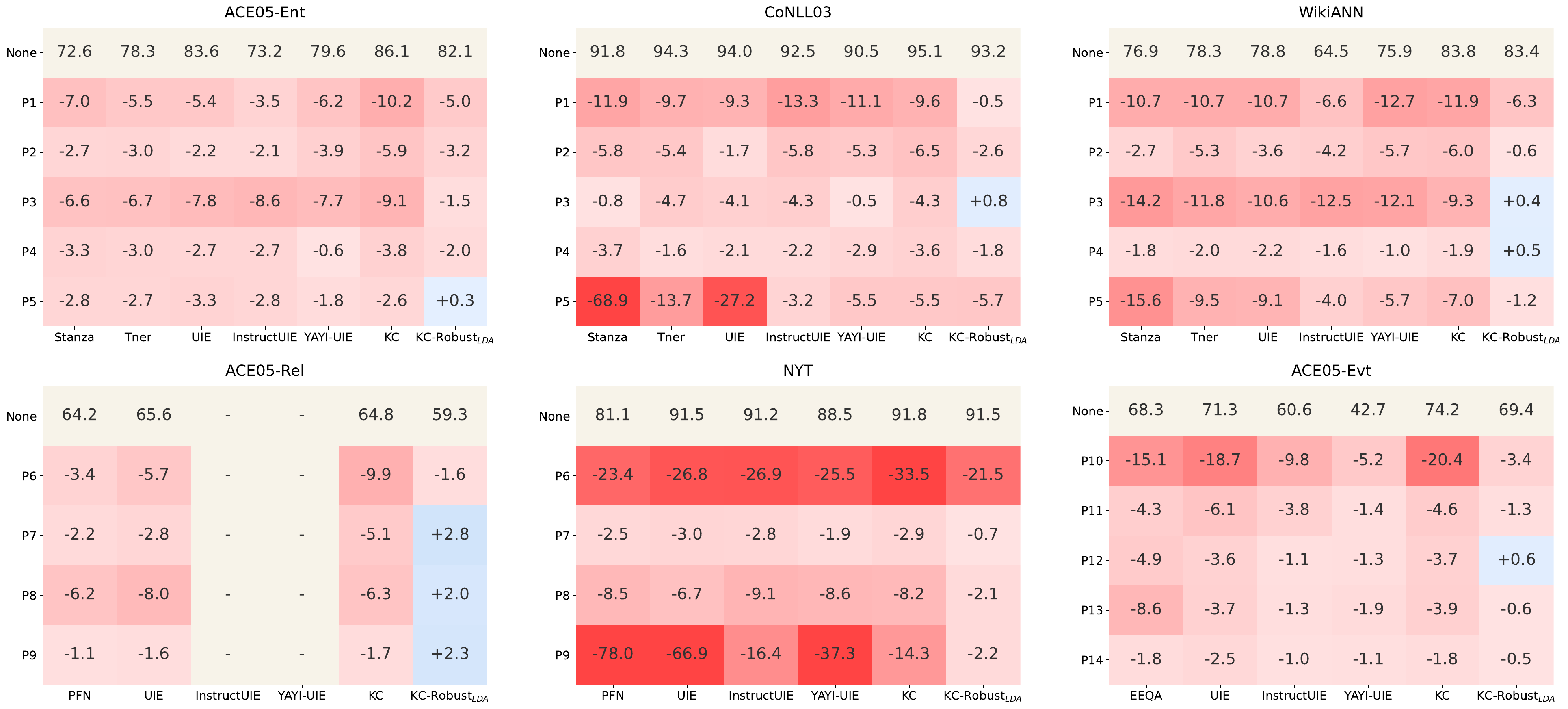}
 \caption{Performance comparison of different models under various perturbations on different datasets. Red and blue indicate performance drop and improvement, respectively
 . KC is short for the KnowCoder model. }
 \vspace{-4mm}
  \label{ner_hotmap}
\end{figure*}

To further verify the effect of different perturbations on the traditional models and UIE models, we report the performance drop and improvement under various perturbations on the six datasets in RUIE-Bench.  
Detailed results are shown in Figure~\ref{ner_hotmap}. Through analysis, we can summarize the following important observations.

(1) Under the three different perturbations of Replace Entity(P1), Replace Triple(P6), and Replace Trigger(P10) to the original test set results in a significant drop in model performance. Therefore, to a certain extent, it shows that the original model may have memorized some specific patterns instead of reasoning based on context. A specific case of a Replace Entity is provided in Appendix~\ref{appendix:prompt7}.

(2) Under certain perturbation settings, such as Lowercase Conversion, we observe that the model's performance drops significantly on some datasets while other datasets remain unaffected. This is because the annotations in the affected datasets contain a relatively high number of uppercase characters. Additionally, we find that the models using LLMs show better performance on these datasets. This suggests that LLMs already have strong generalization abilities to handle such simple noise.

(3)  $\text{KnowCoder-7B-Robust}_{\text{LDA}}$ shows a remarkable improvement across most of the datasets under nearly all perturbations. This observation strongly supports the effectiveness and feasibility of the LDA strategy. Furthermore, an interesting finding is observed in ACE05-Ent, WikiANN, and ACE05-Evt: some models without data augmentation training (such as InstructUIE and YAYI-UIE) show similar robustness to $\text{KnowCoder-7B-Robust}_{\text{LDA}}$. This is because the performance of these models on the original test set is relatively poor, and thus, the perturbations have little effect on their performance.

\section{Conclusion}

In this paper, we introduced RUIE-Bench, a benchmark dataset designed to evaluate the robustness of UIE models. The dataset includes 14 adversarial perturbations for three core IE tasks, i.e., NER, RE, and ED. Through comprehensively benchmarking of existing models, the results reveal that these models struggle to maintain robustness when faced with these adversarial perturbations, highlighting the urgent need for robustness improvement for UIE. Motivated by this, we proposed a Loss-guided Data Augmentation (LDA) method that iteratively selects challenging samples for training. The results demonstrate that LDA achieves performance comparable to fully trained models on RUIE-Bench and even exhibits superior generalization capabilities on unseen datasets. This work aims to provide a valuable benchmark for evaluating robustness in UIE tasks and offer a practical methodology for enhancing model robustness.

\section*{Limitations}
Generating more realistic perturbations remains an exploratory direction for future work. Although we propose various perturbation generation methods for UIE, they still fail to cover the diverse noise present in real-world scenarios. Meanwhile, due to cost and resource constraints, we have not conducted robustness evaluations on more LLMs. Moreover, the performance improvement achieved by the loss-guided data augmentation method may be constrained by the quality of the augmented data. Addressing these issues will be a priority for future work.

\bibliography{custom}

\appendix

\label{sec:appendix}

\section{Prompts for Adversarial Example Generation}
\label{appendix:prompt1}

\begin{figure*}[tbp]  
 \centering
 \includegraphics[width=1.0\textwidth]{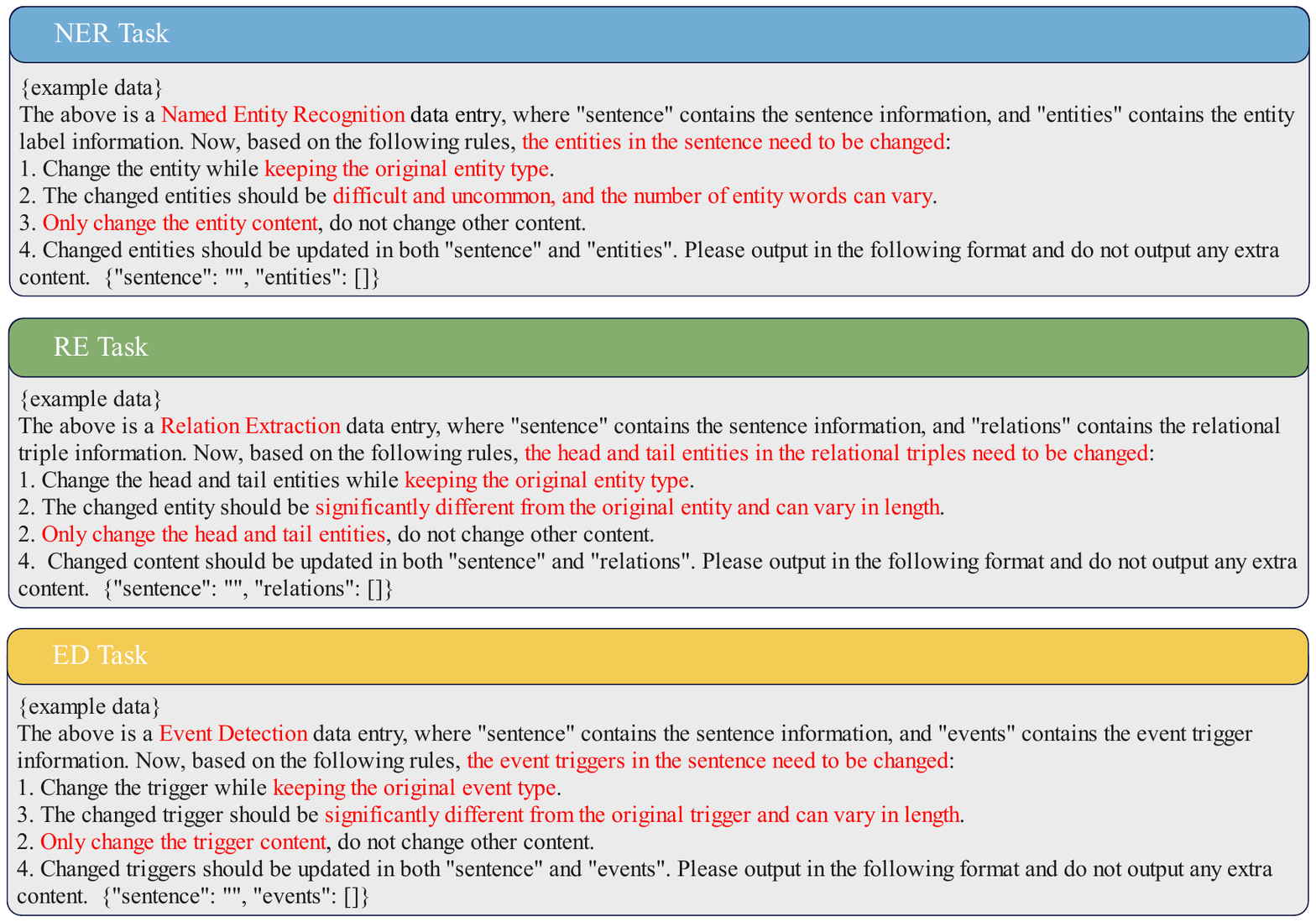}
 \caption{Prompts for Replace Entity, Triple, and Trigger.}
 \vspace{-4mm}
  \label{fig:replace}
\end{figure*}

\paragraph*{Replace Entity, Triple, and Trigger}
In order to generate high-quality adversarial examples, we replace the entities, relation triples, and event triggers in the samples based on some rules for LLMs. The prompts are shown in Figure ~\ref{fig:replace}.

\begin{figure*}[tbp]  
 \centering
 \includegraphics[width=1.0\textwidth]{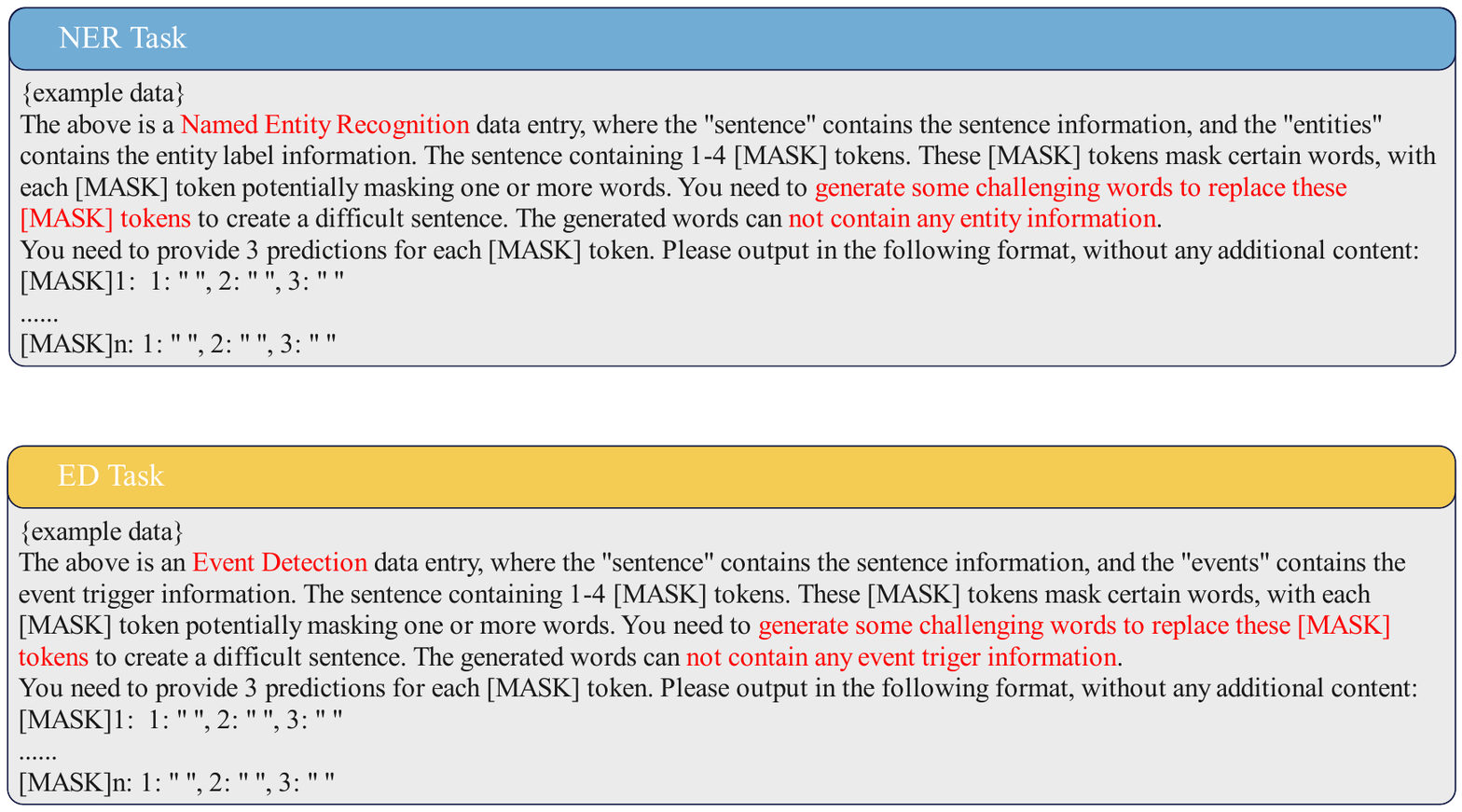}
 \caption{Prompts for Change Context.}
 \vspace{-4mm}
  \label{fig:context}
\end{figure*}

\paragraph*{Change Context}
We change the context of NER and ED samples based on the rules mentioned in Section~\ref{perturbation methods} to generate high-quality adversarial examples. The prompt is shown in Figure ~\ref{fig:context}.

\begin{figure*}[tbp]  
 \centering
 \includegraphics[width=1.0\textwidth]{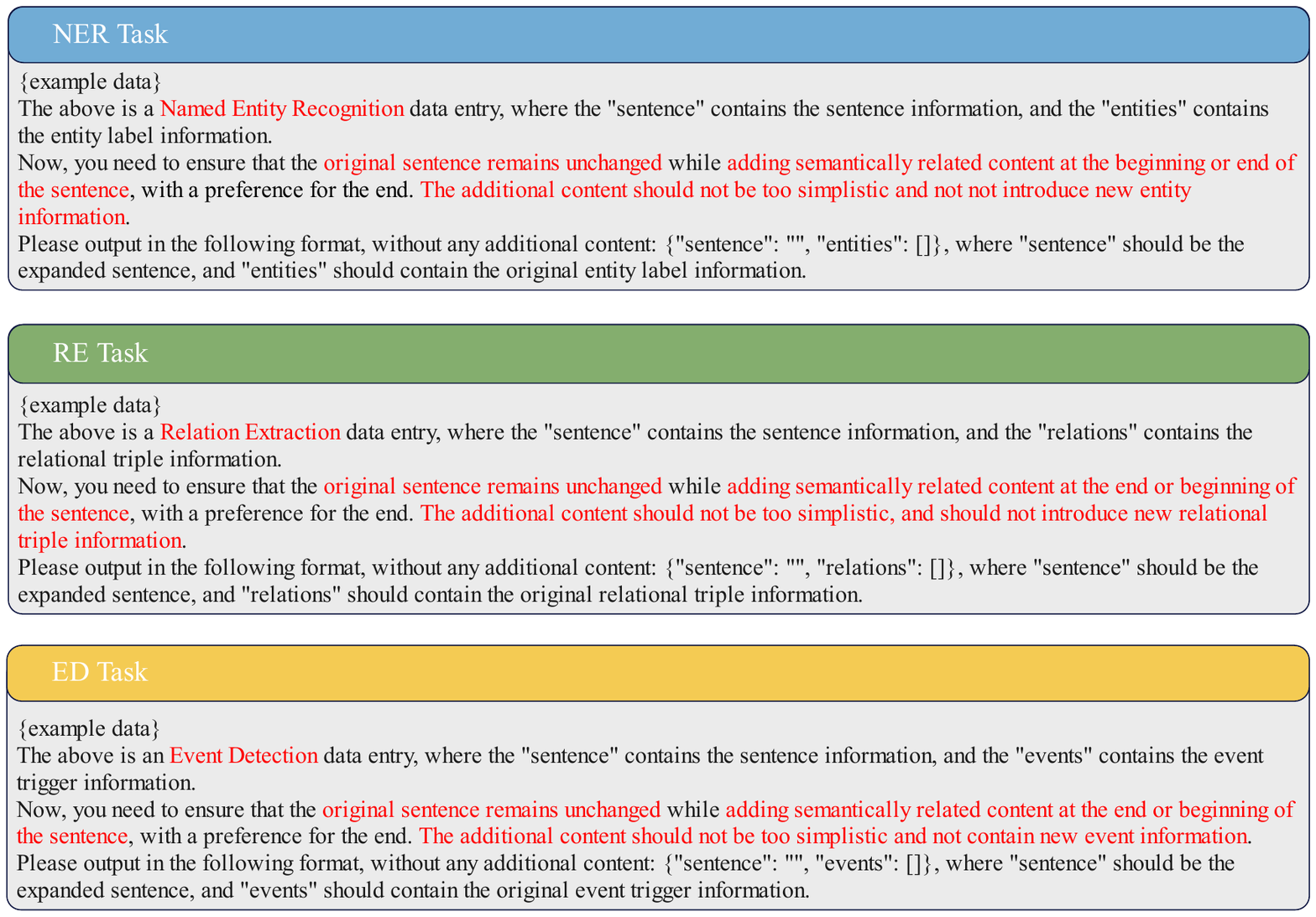}
 \caption{Prompts for Extend Sentence.}
 \vspace{-4mm}
  \label{fig:extend}
\end{figure*}

\paragraph*{Extend Sentence}
We use the prompt in Figure ~\ref{fig:extend} to generate the extended versions of NER, RE, and ED samples.

\section{RUIE-Bench Details}
\label{appendix:prompt2}
\paragraph*{Sampling Details}
Considering that generating adversarial examples for all test sets of NER and RE tasks would result in a large number of samples and costs, as well as significant evaluation expenses, we conduct sampling on the test sets of the selected datasets to balance evaluation costs and accuracy. The main principle we follow is to ensure that the sampled subsets maintain the same distribution. Specifically, for NER, we select 1,000 samples, including 134 from ACE05-Ent, 294 from CoNLL03, and 572 from WikiANN. For RE, we choose 800 samples, with 230 from ACE05-Rel and 570 from NYT. Due to the small quantity and size of ED dataset, we use the complete dataset to generate adversarial examples. 

\paragraph*{Statistics}
Based on the sampled data, we utilize the perturbation methods described in Section~\ref{perturbation methods} to generate adversarial examples, thereby constructing the RUIE Bench dataset. Table~\ref{RUIE-Bench dataset} provides detailed statistics of the dataset.

\begin{table}[htbp]
 \centering
 \resizebox{1\linewidth}{!}{
 \begin{tabular}{c|c|cc|ccc}
 \toprule  
 \multirow{2}{*}{\textbf{Task}} & \multirow{2}{*}{\textbf{Dataset}} & \multicolumn{2}{c}{\textbf{Original Data}} & \multicolumn{3}{|c}{\textbf{RUIE-Bench Data}} \\
 \cmidrule{3-7}
    & & \textbf{Type}  & \textbf{Test size} & \textbf{Type} & \textbf{Sampling size} & \textbf{Data size} \\
 \midrule
 \multirow{3}{*}{NER} & ACE05 & 7 & 1060  & 7 & 134 & 670 \\
 ~ & CoNLL03 & 4 & 3453 & 4 & 294 & 1470 \\
 ~ & WikiANN & 3 & 10000 & 3 & 572 & 2860 \\
 \midrule
 \multirow{2}{*}{RE} & ACE05 & 6 & 2050 & 6 & 230 & 920 \\
 ~ & NYT & 24 & 5000 & 24 & 570 & 2280 \\
 \midrule
 \multirow{1}{*}{ED} & ACE05 & 33 & 676 & 33 & 676 & 3380 \\
 \bottomrule
 \end{tabular}
 }
 \caption{\label{RUIE-Bench dataset}
 Statistics of the RUIE-Bench dataset.}
\end{table}

\section{Augment Data Generation}
\label{appendix:3}
We use the original training sets of all datasets that are constructed for RUIE-Bench to build augmented data. Given the large scale of these training sets and the variety of perturbations, we randomly sample 30\% of the training data from each dataset. Based on this data, we apply all perturbation injection methods from RUIE-Bench to generate augmented data. To generate a large volume of data while avoiding high costs, we choose Deepseek-V3~\cite{deepseekv3} instead of GPT-4~\cite{GPT4} for data generation. The statistics for all datasets are shown in the Figure~\ref{augemnt}.

\begin{table}[htbp]
 \centering
 \resizebox{0.8\linewidth}{!}{
 \begin{tabular}{c|c|cc}
 \toprule  
 \multirow{2}{*}{\textbf{Task}} & \multirow{2}{*}{\textbf{Dataset}} & \multicolumn{2}{c}{\textbf{Data size}} \\
 \cmidrule{3-4}
    & & \textbf{Original}  & \textbf{Augment} \\
 \midrule
 \multirow{3}{*}{NER} & ACE05 & 7299 & 1,0948 \\
 ~ & CoNLL03 & 14041 & 21061 \\
 ~ & WikiANN & 20000 & 30,000 \\
 \midrule
 \multirow{2}{*}{RE} & ACE05 & 10051 & 12061 \\
 ~ & NYT & 56196 & 67435 \\
 \midrule
 \multirow{1}{*}{ED} & ACE05 & 19216 & 28824 \\
 \bottomrule
 \end{tabular}
 }
 \caption{\label{augemnt}
 Statistics of Augmented Data.}
\end{table}

\section{Few-shot Prompts for UIE}
\label{appendix:5}

Taking NER task as an example, we introduce the prompts used for extraction, as shown in Table~\ref{Few-Shot NER}. The prompt mainly consists of five parts: 1 Task Objective, which indicates the goal of the task. 2 Entity Types, defining the types and descriptions of entities to be extracted. 3 Output Formatting, indicating the format of the output. 4 Examples, providing demonstrations for extraction examples, which are randomly selected from the training set. 5 Current Task, the sentence to be extracted. RE and ED tasks also adopt a similar format, with differences only in types and output formats, as shown in Table~\ref{Few-Shot RE} and Table~\ref{Few-Shot ED} respectively.

\begin{table}[htbp]
 \centering
 \small  
 \setlength\tabcolsep{1pt}  
 \resizebox{1\linewidth}{!}{
 \begin{tabular}{@{}l|c|c|cccc@{}}  
 \toprule
 \multirow{2}{*}{\textbf{Task}} & \multirow{2}{*}{\textbf{Dataset}} & \multirow{2}{*}{\textbf{Perturbation Type}} & \multicolumn{3}{c}{\textbf{Model}} \\
 \cmidrule{4-6}
    & & & \textbf{$\text{KnowCoder}_{\text{partial}}$} & \textbf{$\text{KnowCoder-Robust}$} & \textbf{$\text{KnowCoder-Robust}_{\text{LDA}}$}\\
 \midrule
 \multirow{18}{*}{NER} & \multirow{6}{*}{ACE05} & None & 79.2 & \textbf{82.5} & 82.1 \\
    & & Replace Entity & 72.3 & \textbf{78.9} & 76.3 \\
    & & Change Context & 74.2 & \textbf{80.6} & 78.9 \\
    & & Extend Sentence & 72.1 & \textbf{82.7} & 81.4 \\
    & & Typo Injection & 76.8 & \textbf{82.7} & 80.1 \\
    & & Lowercase Conversion & 77.0 & 82.0 & \textbf{82.4} \\
 \cmidrule{2-6}
    & \multirow{6}{*}{Conll03} & None & 91.7 & 93.0 & \textbf{93.2} \\
    & & Replace Entity & 80.5 & 90.9 & \textbf{92.7} \\
    & & Change Context & 87.2 & 90.2 & \textbf{90.6}\\
    & & Extend Sentence & 91.0 & 93.0 & \textbf{94.0}\\
    & & Typo Injection & 90.0 & \textbf{92.5} & 91.4 \\
    & & Lowercase Conversion & 85.1 & \textbf{89.7} & 87.5 \\
 \cmidrule{2-6}
    & \multirow{6}{*}{WikiANN} & None & 81.9 & 83.1 & \textbf{83.4} \\
    & & Replace Entity & 70.8 & \textbf{76.9} & 76.5 \\
    & & Change Context & 77.8 & 80.8 & \textbf{83.4} \\
    & & Extend Sentence & 78.2 & \textbf{83.9} & 83.8 \\
    & & Typo Injection & 79.4 & 83.6 & \textbf{83.9} \\
    & & Lowercase Conversion & 76.4 & \textbf{82.6} & 82.2 \\
 \midrule
 \multirow{10}{*}{RE} & \multirow{5}{*}{ACE05} & None & 58.6 & \textbf{63.5} & 59.3 \\
    & & Replace Triple & 54.9 & \textbf{61.6} & 57.7 \\
    & & Extend Sentence & 54.1 & \textbf{62.3} & 62.1\\
    & & Typo Injection & 53.9 & 59.7 & \textbf{61.3} \\
    & & Lowercase Conversion & 56.5 & 56.9 & \textbf{61.6}\\
 \cmidrule{2-6}
    & \multirow{5}{*}{NYT} & None & 90.6 & 91.1 & \textbf{91.5} \\
    & & Replace Triple & 62.9 & 67.9 & \textbf{70.0} \\
    & & Extend Sentence & 89.2 & \textbf{91.2} & 90.8\\
    & & Typo Injection & 82.8 & \textbf{89.8} & 89.4 \\
    & & Lowercase Conversion & 51.3 & 89.0 & \textbf{89.3} \\
 \midrule
 \multirow{6}{*}{ED} & \multirow{6}{*}{ACE05} & None & 69.1 & \textbf{70.2} & 69.4\\
    & & Replace Trigger & 55.5 & 65.7 & \textbf{66.0} \\
    & & Change Context & 66.1 & 67.1 & \textbf{68.1}\\
    & & Extend Sentence & 65.8 & \textbf{70.5} & 70.0 \\
    & & Typo Injection & 66.8 & 68.3 & \textbf{68.8} \\
    & & Lowercase Conversion & 67.8 & \textbf{69.6} & 68.9\\
 \midrule
 All & All & \textbf{Average} & 71.8 & \textbf{77.2} & \textbf{77.2} \\
 \bottomrule
 \end{tabular}
 } \caption{Comparison of $\text{KnowCoder}_{\text{partial}}$ and different data augmentation training models on RUIE-Bench.}
  \label{adv_com}
\end{table}

\section{Comparison of Data Augmentation Training Models}
\label{appendix:6}

To verify the performance of the model trained using the LDA strategy, we comprehensively compare it with the non-augmented training model $\text{KnowCoder}_{\text{partial}}$ and the model trained with full augmented data, as shown in Table~\ref{adv_com}, where the optimal performance under each specific perturbation for each dataset is bolded. It can be observed that the best performance consistently occurs in the models that have undergone data augmentation training. Moreover, $\text{KnowCoder-Robust}_{\text{LDA}}$ demonstrates performance comparable to that of $\text{KnowCoder-Robust}$, which proves that the LDA strategy can achieve efficient data augmentation training.

\begin{figure*}[tbp]  
 \centering
 \includegraphics[width=1.0\textwidth]{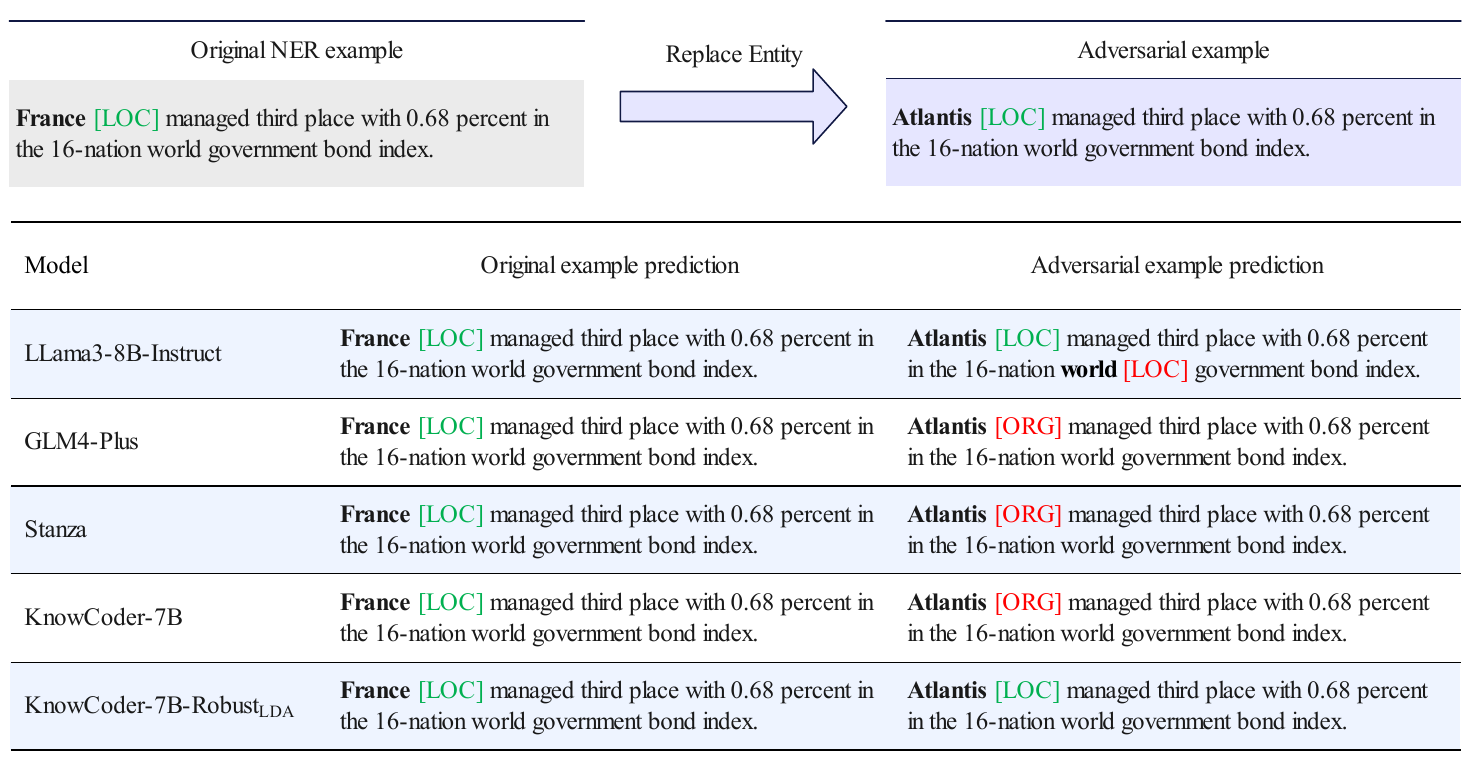}
 \caption{Example cases for the Replace Entity.}
 \vspace{-4mm}
  \label{fig:case_study}
\end{figure*}

\section{Case Study}
\label{appendix:prompt7}
As shown in Figure~\ref{fig:case_study}, we present an example of perturbation on NER. In this example, the correct entity should be relatively easy to identify based on contextual information. For instance, the ``16 - nation world government bond index'' in the sentence is a clear clue. However, in actual prediction, most models make incorrect predictions on the adversarial example, only the model with data augmentation training make correct predictions. This suggests that these models tend to memorize entities rather than using context for inference.

\input{few_shot_ner}

\input{few_shot_re}

\input{few_shot_ed}

\end{document}

%% file: main_table.tex
\begin{table*}
\centering
\adjustbox{max width=\linewidth}{
\begin{tabular}{l|rrrrrrr|rrrrrr|rrrrrrr|rr}
\toprule
\multirow{2}{*}{\textbf{Model}} & \multicolumn{7}{c}{\textbf{\cellcolor{teal!20} NER}} & \multicolumn{6}{c}{\textbf{\cellcolor{red!20} RE}} & \multicolumn{7}{c}{\textbf{\cellcolor{pink!20} ED}} & \multicolumn{2}{c}{\textbf{\cellcolor{blue!10} Overall}} \\ 
\cmidrule(lr){2-8} \cmidrule(lr){9-14} \cmidrule(lr){15-21} \cmidrule(lr){22-23}
& \cellcolor{teal!20}None & \cellcolor{teal!20}P1 & \cellcolor{teal!20}P2 & \cellcolor{teal!20}P3 & \cellcolor{teal!20}P4 & \cellcolor{teal!20}P5 & \cellcolor{teal!20}$\text{Drop}_\text{avg}$ & \cellcolor{red!20}None & \cellcolor{red!20}P6 &\cellcolor{red!20} P7 & \cellcolor{red!20}P8 & \cellcolor{red!20}P9 & \cellcolor{red!20}$\text{Drop}_\text{avg}$ & \cellcolor{pink!20}None & \cellcolor{pink!20}P10 & \cellcolor{pink!20}P11 & \cellcolor{pink!20}P12 & \cellcolor{pink!20}P13 & \cellcolor{pink!20}P14 & \cellcolor{pink!20}$\text{Drop}_\text{avg}$ & \cellcolor{blue!10}Avg & \cellcolor{blue!10}Rank \\ 
\midrule
\multicolumn{23}{l}{\textbf{Open-source LLMs}}\\
\cellcolor{gray!10} Qwen2.5-14B-Instruct & 
\cellcolor{teal!20}58.6 & \cellcolor{teal!20}53.6 & \cellcolor{teal!20}57.7 & \cellcolor{teal!20}55.0 & \cellcolor{teal!20}56.9 & \cellcolor{teal!20}46.6 & \cellcolor{teal!20}$\text{7.9\%}_\downarrow$ & 
\cellcolor{red!20}22.6 & \cellcolor{red!20}19.2 & \cellcolor{red!20}21.3 & \cellcolor{red!20}17.1 & \cellcolor{red!20}8.8 & \cellcolor{red!20}$\text{26.5\%}_\downarrow$ & 
\cellcolor{pink!20}32.0 & \cellcolor{pink!20}29.8 & \cellcolor{pink!20}30.4 & \cellcolor{pink!20}30.9 & \cellcolor{pink!20}31.4 & \cellcolor{pink!20}32.0  & \cellcolor{pink!20}$\text{3.4\%}_\downarrow$ & 
\cellcolor{blue!10}35.5 & \cellcolor{blue!10}11 \\
\cellcolor{gray!10} Qwen2.5-7B-Instruct & 
\cellcolor{teal!20}53.3 & \cellcolor{teal!20}49.8 & \cellcolor{teal!20}51.2 & \cellcolor{teal!20}50.5 & \cellcolor{teal!20}51.2 & \cellcolor{teal!20}41.3 & \cellcolor{teal!20}$\text{8.4\%}_\downarrow$ & 
\cellcolor{red!20}15.6 & \cellcolor{red!20}13.4 & \cellcolor{red!20}14.0 & \cellcolor{red!20}13.8 & \cellcolor{red!20}3.8 & \cellcolor{red!20}$\text{27.9\%}_\downarrow$ & 
\cellcolor{pink!20}19.0 & \cellcolor{pink!20}18.8 & \cellcolor{pink!20}17.6 & \cellcolor{pink!20}17.9 & \cellcolor{pink!20}18.8 & \cellcolor{pink!20}19.2  & \cellcolor{pink!20}$\text{2.8\%}_\downarrow$ & 
\cellcolor{blue!10}27.6 & \cellcolor{blue!10}13 \\
\cellcolor{gray!10} Qwen2.5-3B-Instruct & 
\cellcolor{teal!20}49.5 & \cellcolor{teal!20}47.5 & \cellcolor{teal!20}46.7 & \cellcolor{teal!20}45.3 & \cellcolor{teal!20}45.5 & \cellcolor{teal!20}40.2 & \cellcolor{teal!20}$\text{9.0\%}_\downarrow$ & 
\cellcolor{red!20}8.9 & \cellcolor{red!20}7.6 & \cellcolor{red!20}8.6 & \cellcolor{red!20}7.4 & \cellcolor{red!20}2.0 & \cellcolor{red!20}$\text{28.1\%}_\downarrow$ & 
\cellcolor{pink!20}13.3 & \cellcolor{pink!20}13.8 & \cellcolor{pink!20}12.3 & \cellcolor{pink!20}12.6 & \cellcolor{pink!20}12.7 & \cellcolor{pink!20}13.6  & \cellcolor{pink!20}$\text{2.3\%}_\downarrow$ & 
\cellcolor{blue!10}22.8 & \cellcolor{blue!10}15 \\
\cellcolor{gray!10} Llama3-8B-Instruct & 
\cellcolor{teal!20}55.4 & \cellcolor{teal!20}52.6 & \cellcolor{teal!20}52.9 & \cellcolor{teal!20}51.1 & \cellcolor{teal!20}53.5 & \cellcolor{teal!20}25.7 & \cellcolor{teal!20}$\text{14.9\%}_\downarrow$ & 
\cellcolor{red!20}17.3 & \cellcolor{red!20}15.0 & \cellcolor{red!20}15.7 & \cellcolor{red!20}13.6 & \cellcolor{red!20}2.5 & \cellcolor{red!20}$\text{32.4\%}_\downarrow$ & 
\cellcolor{pink!20}12.8 & \cellcolor{pink!20}13.1 & \cellcolor{pink!20}13.1 & \cellcolor{pink!20}10.9 & \cellcolor{pink!20}12.5 & \cellcolor{pink!20}12.2 & \cellcolor{pink!20}$\text{3.4\%}_\downarrow$ & 
\cellcolor{blue!10}25.3 & \cellcolor{blue!10}14 \\
\cellcolor{gray!10} Glm-4-9B-Chat & 
\cellcolor{teal!20}57.4 & \cellcolor{teal!20}54.0 & \cellcolor{teal!20}55.8 & \cellcolor{teal!20}51.4 & \cellcolor{teal!20}56.6 & \cellcolor{teal!20}43.2 & \cellcolor{teal!20}$\text{9.0\%}_\downarrow$ & 
\cellcolor{red!20}8.8 & \cellcolor{red!20}7.5 & \cellcolor{red!20}7.5 & \cellcolor{red!20}7.4 & \cellcolor{red!20}1.8 & \cellcolor{red!20}$\text{31.2\%}_\downarrow$ & 
\cellcolor{pink!20}5.6 & \cellcolor{pink!20}6.6 & \cellcolor{pink!20}4.7 & \cellcolor{pink!20}4.1 & \cellcolor{pink!20}5.1 & \cellcolor{pink!20}5.9 & \cellcolor{pink!20}$\text{5.7\%}_\downarrow$ & 
\cellcolor{blue!10}22.6 & \cellcolor{blue!10}16 \\
\cellcolor{gray!10} Internlm2.5-7B-Chat & 
\cellcolor{teal!20}51.6 & \cellcolor{teal!20}48.0 & \cellcolor{teal!20}48.8 & \cellcolor{teal!20}46.9 & \cellcolor{teal!20}45.3 & \cellcolor{teal!20}31.0 & \cellcolor{teal!20}$\text{14.7\%}_\downarrow$ & 
\cellcolor{red!20}12.0 & \cellcolor{red!20}11.3 & \cellcolor{red!20}10.1 & \cellcolor{red!20}9.0 & \cellcolor{red!20}1.7 & \cellcolor{red!20}$\text{33.1\%}_\downarrow$ & 
\cellcolor{pink!20}11.0 & \cellcolor{pink!20}10.3 & \cellcolor{pink!20}10.6 & \cellcolor{pink!20}8.2 & \cellcolor{pink!20}9.6 & \cellcolor{pink!20}11.3 & \cellcolor{pink!20}$\text{9.1\%}_\downarrow$ & 
\cellcolor{blue!10}22.2 & \cellcolor{blue!10}17 \\
\cellcolor{gray!10} CodeLlama-7B-Instruct & 
\cellcolor{teal!20}46.3 & \cellcolor{teal!20}45.0 & \cellcolor{teal!20}45.0 & \cellcolor{teal!20}38.9 & \cellcolor{teal!20}42.4 & \cellcolor{teal!20}14.5 & \cellcolor{teal!20}$\text{19.7\%}_\downarrow$ & 
\cellcolor{red!20}13.7 & \cellcolor{red!20}11.6 & \cellcolor{red!20}12.2 & \cellcolor{red!20}11.3 & \cellcolor{red!20}2.8 & \cellcolor{red!20}$\text{30.8\%}_\downarrow$ & 
\cellcolor{pink!20}8.6 & \cellcolor{pink!20}9.3 & \cellcolor{pink!20}8.8 & \cellcolor{pink!20}6.1 & \cellcolor{pink!20}8.2 & \cellcolor{pink!20}9.2 & \cellcolor{pink!20}$\text{3.3\%}_\downarrow$ & 
\cellcolor{blue!10}19.6 & \cellcolor{blue!10}18 \\
\cellcolor{gray!10} Vicuna-7B-v1.5 & 
\cellcolor{teal!20}39.0 & \cellcolor{teal!20}38.2 & \cellcolor{teal!20}37.4 & \cellcolor{teal!20}35.0 & \cellcolor{teal!20}38.0 & \cellcolor{teal!20}16.7 & \cellcolor{teal!20}$\text{15.2\%}_\downarrow$ & 
\cellcolor{red!20}11.2 & \cellcolor{red!20}11.0 & \cellcolor{red!20}10.1 & \cellcolor{red!20}7.6 & \cellcolor{red!20}0.8 & \cellcolor{red!20}$\text{34.1\%}_\downarrow$ & 
\cellcolor{pink!20}6.9 & \cellcolor{pink!20}7.5 & \cellcolor{pink!20}7.2 & \cellcolor{pink!20}4.5 & \cellcolor{pink!20}6.1 & \cellcolor{pink!20}6.3 & \cellcolor{pink!20}$\text{8.4\%}_\downarrow$ & 
\cellcolor{blue!10}16.7 & \cellcolor{blue!10}19 \\

\multicolumn{23}{l}{\textbf{Closed-source LLMs}}\\
\cellcolor{gray!10} GLM4-Plus & 
\cellcolor{teal!20}63.2 & \cellcolor{teal!20}59.8 & \cellcolor{teal!20}63.0 & \cellcolor{teal!20}61.6 & \cellcolor{teal!20}60.9 & \cellcolor{teal!20}49.7 & \cellcolor{teal!20}$\text{6.6\%}_\downarrow$ & 
\cellcolor{red!20}32.2 & \cellcolor{red!20}29.2 & \cellcolor{red!20}31.3 & \cellcolor{red!20}26.1 & \cellcolor{red!20}5.3 & \cellcolor{red!20}$\text{28.6\%}_\downarrow$ & 
\cellcolor{pink!20}43.5 & \cellcolor{pink!20}39.9 & \cellcolor{pink!20}43.3 & \cellcolor{pink!20}34.6 & \cellcolor{pink!20}40.8 & \cellcolor{pink!20}43.9 & \cellcolor{pink!20}$\text{6.9\%}_\downarrow$ & 
\cellcolor{blue!10}42.8 & \cellcolor{blue!10}8 \\
\cellcolor{gray!10} DeepSeek-V3 & 
\cellcolor{teal!20}62.3 & \cellcolor{teal!20}59.8 & \cellcolor{teal!20}61.5 & \cellcolor{teal!20}61.3 & \cellcolor{teal!20}58.7 & \cellcolor{teal!20}55.0 & \cellcolor{teal!20}$\text{4.9\%}_\downarrow$ & 
\cellcolor{red!20}31.3 & \cellcolor{red!20}29.0 & \cellcolor{red!20}29.6 & \cellcolor{red!20}26.2 & \cellcolor{red!20}10.0 & \cellcolor{red!20}$\text{24.3\%}_\downarrow$ & 
\cellcolor{pink!20}38.8 & \cellcolor{pink!20}37.8 & \cellcolor{pink!20}38.3 & \cellcolor{pink!20}34.5 & \cellcolor{pink!20}35.6 & \cellcolor{pink!20}38.9 & \cellcolor{pink!20}$\text{4.6\%}_\downarrow$  & 
\cellcolor{blue!10}41.7 & \cellcolor{blue!10}9\\
\cellcolor{gray!10} GPT-4-turbo & 
\cellcolor{teal!20}60.6 & \cellcolor{teal!20}57.5 & \cellcolor{teal!20}59.8 & \cellcolor{teal!20}58.2 & \cellcolor{teal!20}56.2 & \cellcolor{teal!20}33.4 & \cellcolor{teal!20}$\text{12.5\%}_\downarrow$ & 
\cellcolor{red!20}33.0 & \cellcolor{red!20}30.0 & \cellcolor{red!20}31.6 & \cellcolor{red!20}26.8 & \cellcolor{red!20}4.5 & \cellcolor{red!20}$\text{29.6\%}_\downarrow$ & 
\cellcolor{pink!20}40.0  & \cellcolor{pink!20}38.0  & \cellcolor{pink!20}39.6  & \cellcolor{pink!20}34.5  & \cellcolor{pink!20}37.3  & \cellcolor{pink!20}39.8  & \cellcolor{pink!20}$\text{5.4\%}_\downarrow$  &
\cellcolor{blue!10}40.0 & \cellcolor{blue!10}10 \\
\cellcolor{gray!10} GPT-3.5-turbo & 
\cellcolor{teal!20}51.8 & \cellcolor{teal!20}47.9 & \cellcolor{teal!20}48.9 & \cellcolor{teal!20}50.5 & \cellcolor{teal!20}39.0 & \cellcolor{teal!20}33.1 & \cellcolor{teal!20}$\text{15.3\%}_\downarrow$ & 
\cellcolor{red!20}23.8 & \cellcolor{red!20}20.6 & \cellcolor{red!20}21.3 & \cellcolor{red!20}16.7 & \cellcolor{red!20}2.4 & \cellcolor{red!20}$\text{35.9\%}_\downarrow$ & 
\cellcolor{pink!20}38.0 & \cellcolor{pink!20}29.5 & \cellcolor{pink!20}36.1 & \cellcolor{pink!20}36.7 & \cellcolor{pink!20}33.9 & \cellcolor{pink!20}36.9 & \cellcolor{pink!20}$\text{8.9\%}_\downarrow$ & 
\cellcolor{blue!10}33.3 & \cellcolor{blue!10}12 \\
\midrule
\multicolumn{23}{l}{\textbf{Traditional IE Models}}\\
\cellcolor{gray!10} Stanza & 
\cellcolor{teal!20}80.7 & \cellcolor{teal!20}70.1 & \cellcolor{teal!20}77.1 & \cellcolor{teal!20}71.5 & \cellcolor{teal!20}78.1 & \cellcolor{teal!20}51.1 & \cellcolor{teal!20}$\text{13.8\%}_\downarrow$ & 
\cellcolor{red!20}- & \cellcolor{red!20}- & \cellcolor{red!20}- & \cellcolor{red!20}- & \cellcolor{red!20}- & \cellcolor{red!20}- & 
\cellcolor{pink!20}-  & \cellcolor{pink!20}-  & \cellcolor{pink!20}-  & \cellcolor{pink!20}-  & \cellcolor{pink!20}-  & \cellcolor{pink!20}-  & \cellcolor{pink!20}-  &
\cellcolor{blue!10}- & \cellcolor{blue!10}-\\
\cellcolor{gray!10} TNER & 
\cellcolor{teal!20}83.0 & \cellcolor{teal!20}73.3 & \cellcolor{teal!20}78.0 & \cellcolor{teal!20}73.9 & \cellcolor{teal!20}81.0 & \cellcolor{teal!20}73.2 & \cellcolor{teal!20}$\text{8.6\%}_\downarrow$ & 
\cellcolor{red!20}- & \cellcolor{red!20}- & \cellcolor{red!20}- & \cellcolor{red!20}- & \cellcolor{red!20}- & \cellcolor{red!20}- & 
\cellcolor{pink!20}-  & \cellcolor{pink!20}-  & \cellcolor{pink!20}-  & \cellcolor{pink!20}-  & \cellcolor{pink!20}-  & \cellcolor{pink!20}-  & \cellcolor{pink!20}-  &
\cellcolor{blue!10}- & \cellcolor{blue!10}- \\
\cellcolor{gray!10} PFN & 
\cellcolor{teal!20}- & \cellcolor{teal!20}- & \cellcolor{teal!20}- & \cellcolor{teal!20}- & \cellcolor{teal!20}- & \cellcolor{teal!20}- & \cellcolor{teal!20}- & 
\cellcolor{red!20}76.3 & \cellcolor{red!20}58.6 & \cellcolor{red!20}73.8 & \cellcolor{red!20}68.4 & \cellcolor{red!20}20.4 & \cellcolor{red!20}$\text{27.5\%}_\downarrow$  & 
\cellcolor{pink!20}-  & \cellcolor{pink!20}-  & \cellcolor{pink!20}-  & \cellcolor{pink!20}-  & \cellcolor{pink!20}-  & \cellcolor{pink!20}-  & \cellcolor{pink!20}-  & 
\cellcolor{blue!10}- & \cellcolor{blue!10}-  \\
\cellcolor{gray!10} EEQA & 
\cellcolor{teal!20}- & \cellcolor{teal!20}- & \cellcolor{teal!20}- & \cellcolor{teal!20}- & \cellcolor{teal!20}- & \cellcolor{teal!20}- & \cellcolor{teal!20}- & 
\cellcolor{red!20}- & \cellcolor{red!20}- & \cellcolor{red!20}- & \cellcolor{red!20}- & \cellcolor{red!20}- & \cellcolor{red!20}- &
\cellcolor{pink!20}68.3  & \cellcolor{pink!20}53.2  & \cellcolor{pink!20}64.0  & \cellcolor{pink!20}63.4  & \cellcolor{pink!20}59.7  & \cellcolor{pink!20}66.5  & \cellcolor{pink!20}$\text{10.1\%}_\downarrow$ & 
\cellcolor{blue!10}- & \cellcolor{blue!10}-  \\
\multicolumn{23}{l}{\textbf{UIE Models}}\\
\cellcolor{gray!10} UIE & 
\cellcolor{teal!20}83.9 & \cellcolor{teal!20}74.3 & \cellcolor{teal!20}81.1 & \cellcolor{teal!20}75.6 & \cellcolor{teal!20}81.7 & \cellcolor{teal!20}70.3 & \cellcolor{teal!20}$\text{8.7\%}_\downarrow$ & 
\cellcolor{red!20}\textbf{84.1} & \cellcolor{red!20}63.3 & \cellcolor{red!20}81.1 & \cellcolor{red!20}77.0 & \cellcolor{red!20}35.9 & \cellcolor{red!20}$\text{23.5\%}_\downarrow$ & 
\cellcolor{pink!20}71.3 & \cellcolor{pink!20}52.6 & \cellcolor{pink!20}65.2 & \cellcolor{pink!20}67.7 & \cellcolor{pink!20}67.6 & \cellcolor{pink!20}68.8 & \cellcolor{pink!20}$\text{9.9\%}_\downarrow$ & 
\cellcolor{blue!10}70.7 & \cellcolor{blue!10}5\\
\cellcolor{gray!10} InstructUIE-11B & 
\cellcolor{teal!20}73.9 & \cellcolor{teal!20}65.7 & \cellcolor{teal!20}69.5 & \cellcolor{teal!20}64.3 & \cellcolor{teal!20}72.0 & \cellcolor{teal!20}70.3 & \cellcolor{teal!20}$\text{7.5\%}_\downarrow$ & 
\cellcolor{red!20}68.4 & \cellcolor{red!20}48.3 & \cellcolor{red!20}66.3 & \cellcolor{red!20}61.6 & \cellcolor{red!20}56.1 & \cellcolor{red!20}$\text{15.1\%}_\downarrow$  & 
\cellcolor{pink!20}60.6  & \cellcolor{pink!20}50.8  & \cellcolor{pink!20}56.8  & \cellcolor{pink!20}59.5  & \cellcolor{pink!20}59.3  & \cellcolor{pink!20}59.6  & \cellcolor{pink!20}$\text{5.6\%}_\downarrow$  & 
\cellcolor{blue!10}62.5 & \cellcolor{blue!10}6  \\
\cellcolor{gray!10} YAYI-UIE-13B & 
\cellcolor{teal!20}80.7 & \cellcolor{teal!20}69.3 & \cellcolor{teal!20}75.3 & \cellcolor{teal!20}72.6 & \cellcolor{teal!20}79.2 & \cellcolor{teal!20}75.6 & \cellcolor{teal!20}$\text{7.8\%}_\downarrow$ & 
\cellcolor{red!20}66.4 & \cellcolor{red!20}47.3 & \cellcolor{red!20}65.0 & \cellcolor{red!20}59.9 & \cellcolor{red!20}38.4 & \cellcolor{red!20}$\text{20.7\%}_\downarrow$  & 
\cellcolor{pink!20}42.7  & \cellcolor{pink!20}37.5  & \cellcolor{pink!20}41.3  & \cellcolor{pink!20}41.4  & \cellcolor{pink!20}40.8  & \cellcolor{pink!20}41.6  & \cellcolor{pink!20}$\text{5.2\%}_\downarrow$  & 
\cellcolor{blue!10}57.4 & \cellcolor{blue!10}7  \\
\cellcolor{gray!10} KnowCoder-7B & 
\cellcolor{teal!20}\textbf{87.4} & \cellcolor{teal!20}76.4 & \cellcolor{teal!20}81.3 & \cellcolor{teal!20}79.6 & \cellcolor{teal!20}84.7 & \cellcolor{teal!20}81.5 & \cellcolor{teal!20}$\text{7.7\%}_\downarrow$ & 
\cellcolor{red!20}84.0 & \cellcolor{red!20}57.3 & \cellcolor{red!20}80.5 & \cellcolor{red!20}76.4 & \cellcolor{red!20}73.3 & \cellcolor{red!20}$\text{14.4\%}_\downarrow$  & 
\cellcolor{pink!20}\textbf{74.2}  & \cellcolor{pink!20}53.8  & \cellcolor{pink!20}\textbf{69.6}  & \cellcolor{pink!20}\textbf{70.5} & \cellcolor{pink!20}\textbf{70.3}  & \cellcolor{pink!20}\textbf{72.4}  & \cellcolor{pink!20}$\text{9.3\%}_\downarrow$  & 
\cellcolor{blue!10}74.9 & \cellcolor{blue!10}3  \\
\cellcolor{gray!10} $\text{KnowCoder-7B}_\text{partial}$ & 
\cellcolor{teal!20}84.4 & \cellcolor{teal!20}73.8 & \cellcolor{teal!20}80.1 & \cellcolor{teal!20}81.1 & \cellcolor{teal!20}82.1 & \cellcolor{teal!20}79.0 & \cellcolor{teal!20}$\text{6.1\%}_\downarrow$ & 
\cellcolor{red!20}81.4 & \cellcolor{red!20}60.6 & \cellcolor{red!20}79.1 & \cellcolor{red!20}74.5 & \cellcolor{red!20}52.8 & \cellcolor{red!20}$\text{18.0\%}_\downarrow$  & 
\cellcolor{pink!20}69.1  & \cellcolor{pink!20}55.5  & \cellcolor{pink!20}66.1  & \cellcolor{pink!20}65.8  & \cellcolor{pink!20}66.8  & \cellcolor{pink!20}67.8  & \cellcolor{pink!20}$\text{6.8\%}_\downarrow$  & 
\cellcolor{blue!10}71.8 & \cellcolor{blue!10}4  \\
\midrule
\cellcolor{gray!10} $\text{KnowCoder-7B-Robust}$ & 
\cellcolor{teal!20}85.9 & \cellcolor{teal!20}\textbf{81.3} & \cellcolor{teal!20}83.5 & \cellcolor{teal!20}86.4 & \cellcolor{teal!20}\textbf{86.1} & \cellcolor{teal!20}\textbf{84.6} & \cellcolor{teal!20}\textbf{$\text{1.7\%}_\downarrow$} & 
\cellcolor{red!20}83.1 & \cellcolor{red!20}66.0 & \cellcolor{red!20}\textbf{82.9} & \cellcolor{red!20}81.1 & \cellcolor{red!20}79.8  & \cellcolor{red!20}$\text{6.8\%}_\downarrow$  & 
\cellcolor{pink!20}70.2 & \cellcolor{pink!20}65.7  & \cellcolor{pink!20}67.1  & \cellcolor{pink!20}\textbf{70.5}  & \cellcolor{pink!20}68.3  & \cellcolor{pink!20}69.6  & \cellcolor{pink!20}$\text{2.8\%}_\downarrow$ & 
\cellcolor{blue!10}\textbf{77.2} & \cellcolor{blue!10}1  \\
\cellcolor{gray!10} $\text{KnowCoder-7B-Robust}_{\text{LDA}}$ & 
\cellcolor{teal!20}86.1 & \cellcolor{teal!20}81.2 & \cellcolor{teal!20}\textbf{84.9} & \cellcolor{teal!20}\textbf{86.5} & \cellcolor{teal!20}85.6 & \cellcolor{teal!20}83.8 & \cellcolor{teal!20}$\text{1.9\%}_\downarrow$ & 
\cellcolor{red!20}82.2 & \cellcolor{red!20}\textbf{66.5} & \cellcolor{red!20}82.5 & \cellcolor{red!20}\textbf{81.3} & \cellcolor{red!20}\textbf{81.3} & \cellcolor{red!20}\textbf{$\text{5.2\%}_\downarrow$}  & 
\cellcolor{pink!20}69.4  & \cellcolor{pink!20}\textbf{66.0}  & \cellcolor{pink!20}68.1  & \cellcolor{pink!20}70.0  & \cellcolor{pink!20}68.8  & \cellcolor{pink!20}68.9  & \cellcolor{pink!20}\textbf{$\text{1.5\%}_\downarrow$}  & 
\cellcolor{blue!10}\textbf{77.2} & \cellcolor{blue!10}1  \\
\bottomrule
\end{tabular}
}
\caption{The performance of all baselines and our models on RUIE-bench.}
\label{tab:robust evaluation}
\end{table*}

%% file: few_shot_ner.tex
\begingroup
\begin{table*}[htp]
    \centering
    
    \begin{tabular}{p{\textwidth}}
        \toprule
        \underline{\textbf{\textsc{Prompt for Few-Shot NER.}}} \\
        \vspace{-2mm}
        \#\# Task Objective \\
        Perform Named Entity Recognition (NER) on input sentences to extract entities of these types:\\ \\
        \#\#Entity Types: \\
        \{entity\_type 1\}: \{description 1\}\\
        \{entity\_type 2\}: \{description 2\}\\
        ······ \\
        \{entity\_type n\}: \{description n\}\\ The entity type here refer to the entity types specific to a given dataset, where the description represents the entity type information. When used, it should be replaced with the actual entity types and corresponding descriptions based on the specific dataset.\\ \\

        \#\# Output Formatting \\
        1. Return a JSON list of entities \\
        2. Each entity must include: \\
        - **Type**: Entity category (exact uppercase labels) \\
        - **Name**: Original text span \\
        3. Return empty list if no entities found \\ \\

        \#\# Examples (10-shot)  \\
        1. Input: \{example sentence 1\}\\
        \hspace{1em}Output: \{recognition result 1\} \\
        2. Input: \{example sentence 2\}\\
        \hspace{1em}Output: \{recognition result 2\} \\
        ······ \\
        10. Input: \{example sentence 10\}\\
        \hspace{1em}Output: \{recognition result 10\} \\ 
        The example sentences are selected from the training set, and the recognition results should fully comply with the defined Output Formatting.\\ \\

        \#\# Current Task \\
        Input: \{test sentence\} \\
        Output: \\
        The input here should be the sentences to be tested, and the output should be the model's recognition results.\\
        

        \bottomrule
    \end{tabular}
    
    \caption{Prompt for Few-Shot NER.}
    \label{Few-Shot NER}
\end{table*}

\endgroup

%% file: few_shot_re.tex
\begingroup
\begin{table*}[htp]
    \centering
    
    \begin{tabular}{p{\textwidth}}
        \toprule
        \underline{\textbf{\textsc{Prompt for Few-Shot RE.}}} \\
        \vspace{-2mm}
        \#\# Task Objective \\
        Perform Relation Extraction (RE) on input sentences to extract relational triples of these types:\\ \\
        \#\#Relation Types: \\
        \{relation\_type 1\}: \{description 1\}\\
        \{relation\_type 2\}: \{description 2\}\\
        ······ \\
        \{relation\_type n\}: \{description n\}\\ The relation type here refer to the relation types specific to a given dataset, where the description represents the relation type information. When used, it should be replaced with the actual relation types and corresponding descriptions based on the specific dataset.\\ \\

        \#\# Output Formatting \\
        1. Return a JSON list of relational triples \\
        2. Each relational triple must include: \\
        - **Head**: Original head entity span \\
        - **Type**: Relation category (exact uppercase labels) \\
        - **Tail**: Original tail entity span \\
        3. Return empty list if no relational triples found \\ \\

        \#\# Examples (10-shot)  \\
        1. Input: \{example sentence 1\}\\
        \hspace{1em}Output: \{recognition result 1\} \\
        2. Input: \{example sentence 2\}\\
        \hspace{1em}Output: \{recognition result 2\} \\
        ······ \\
        10. Input: \{example sentence 10\}\\
        \hspace{1em}Output: \{recognition result 10\} \\ 
        The example sentences are selected from the training set, and the recognition results should fully comply with the defined Output Formatting.\\ \\

        \#\# Current Task \\
        Input: \{test sentence\} \\
        Output: \\
        The input here should be the sentences to be tested, and the output should be the model's recognition results.\\
        

        \bottomrule
    \end{tabular}
    
    \caption{Prompt for Few-Shot RE.}
    \label{Few-Shot RE}
\end{table*}

\endgroup

%% file: few_shot_ed.tex
\begingroup
\begin{table*}[htp]
    \centering
    
    \begin{tabular}{p{\textwidth}}
        \toprule
        \underline{\textbf{\textsc{Prompt for Few-Shot ED.}}} \\
        \vspace{-2mm}
        \#\# Task Objective \\
        Perform Event Detection (ED) on input sentences to extract events of these types:\\ \\
        \#\#Event Types: \\
        \{event\_type 1\}: \{description 1\}\\
        \{event\_type 2\}: \{description 2\}\\
        ······ \\
        \{event\_type n\}: \{description n\}\\ The event type here refer to the event types specific to a given dataset, where the description represents the event type information. When used, it should be replaced with the actual event types and corresponding descriptions based on the specific dataset.\\ \\

        \#\# Output Formatting \\
        1. Return a JSON list of events \\
        2. Each event must include: \\
        - **Type**: Event category (exact uppercase labels) \\
        - **Trigger**: Event trigger span \\
        3. Return empty list if no event found \\ \\

        \#\# Examples (10-shot)  \\
        1. Input: \{example sentence 1\}\\
        \hspace{1em}Output: \{recognition result 1\} \\
        2. Input: \{example sentence 2\}\\
        \hspace{1em}Output: \{recognition result 2\} \\
        ······ \\
        10. Input: \{example sentence 10\}\\
        \hspace{1em}Output: \{recognition result 10\} \\ 
        The example sentences are selected from the training set, and the recognition results should fully comply with the defined Output Formatting.\\ \\

        \#\# Current Task \\
        Input: \{test sentence\} \\
        Output: \\
        The input here should be the sentences to be tested, and the output should be the model's recognition results.\\
        

        \bottomrule
    \end{tabular}
    
    \caption{Prompt for Few-Shot ED.}
    \label{Few-Shot ED}
\end{table*}

\endgroup